\documentclass{article}

\usepackage{microtype}
\usepackage{graphicx}
\usepackage{subfigure}
\usepackage{amsfonts}
\usepackage{multirow}
\usepackage{booktabs} % for professional tables
\usepackage{makecell}
\usepackage{amsmath}

\usepackage{hyperref}

\usepackage[accepted]{icml2021}

\icmltitlerunning{Self Attention with Temporal Prior}

\begin{document}

\twocolumn[
\icmltitle{Self Attention with Temporal Prior: Can We Learn More from Arrow of Time?}



\icmlsetsymbol{equal}{*}

\begin{icmlauthorlist}
\icmlauthor{Kyung Geun Kim}{equal,vuno}
\icmlauthor{Byeong Tak Lee}{equal,medai,loc}
\end{icmlauthorlist}

\icmlaffiliation{vuno}{VUNO Inc.}
\icmlaffiliation{medai}{Medical AI Inc.}
\icmlaffiliation{loc}{This research was conducted while the author was working under VUNO Inc}
\icmlcorrespondingauthor{Kyung Geun Kim}{kyunggk@stanford.edu}

\icmlkeywords{Machine Learning, ICML}

\vskip 0.3in
]



\printAffiliationsAndNotice{\icmlEqualContribution} % otherwise use the standard text.

\begin{abstract}
Many diverse phenomena in nature often inherently encode both short- and long-term temporal dependencies, which especially result from the direction of the flow of time. In this respect, we discovered experimental evidence suggesting that {\it interrelations} of these events are higher for closer time stamps. However, to be able for attention-based models to learn these regularities in short-term dependencies, it requires large amounts of data, which are often infeasible. This is because, while they are good at learning piece-wise temporal dependencies, attention-based models lack structures that encode biases in time series. As a resolution, we propose a simple and efficient method that enables attention layers to better encode the short-term temporal bias of these data sets by applying learnable, adaptive kernels directly to the attention matrices. We chose various prediction tasks for the experiments using Electronic Health Records (EHR) data sets since they are great examples with underlying long- and short-term temporal dependencies. Our experiments show exceptional classification results compared to best-performing models on most tasks and data sets. 
\end{abstract}

\section{Introduction}
\label{intro}
Time series are realizations of diverse phenomena in nature with underlying dynamics of long and short-term temporal dependencies. Although the long-term and point-wise dependencies can also be critical for certain tasks, it is natural to believe that data points that are close in time are highly \textit{interrelational} \cite{tonekaboni2021unsupervised}. One reason this might be plausible is that the direction of time is linear and asymmetric in the physical world. Particular models with such underlying philosophy and inductive bias, like Markov Chains, are useful in many situations. With these observations, we hypothesized that allowing models to directly encode this \textit{interrelational} time dependencies while still attending to long-term time dependencies can introduce a powerful inductive bias for better performance in relevant prediction tasks. 

RNNs, including their extensions, have these inductive biases already encoded \cite{battaglia2018relational}. However, it is well known that, although RNNs can theoretically account for long-term dependencies, learning long-term dependencies with gradient descent is difficult in practice \cite{279181}. Attention-based models, such as Transformers, are designed to solve these issues \cite{vaswani2017attention}. While Transformers, with the proper structure, can attend to any pairs of discrete time stamps in parallel, it has no monotonic, \textit{interrelational} bias between data points that are close in the temporal domain \cite{dosovitskiy2021image}. We hypothesized that by explicitly enabling the capacity to learn these temporal biases, Transformers could learn relevant underlying temporal regularities that are more optimal for prediction tasks of interest. 

To best achieve this, we reasoned that encouraging specific structures on the attention matrices would be the most simple and efficient solution. By designing kernels inspired by various correlation functions applied to the attention matrix directly, we propose an extension to the Transformer with Self Attention with Temporal Prior (SAT-Transformer). The designed kernels are parameterized with minimal amounts of learnable parameters to enforce temporal bias to some degree. Although the main experiments use Electronic Health Records (EHR) data sets, which take the standard form of time series data sets, we believe the SAT-Transformer is generalizable to other time series data sets. The SAT-Transformer achieved significant performance gains compared to the current best-performing model on various prediction tasks across multiple data sets. In addition, we have implemented the kernels of SAT-Transformers in a way that vectorized element-wise matrix multiplication can be possible. Compared to vanilla Transformers, SAT-Transformer only requires minimal additional computation, which can be done efficiently. As a result, SAT-Transformer is far more effective than Transformers and RNNs performance-wise while being almost as efficient as Transformers. 

    \begin{figure*}[h!]
      \centering
        \centerline{\includegraphics[width=17cm]{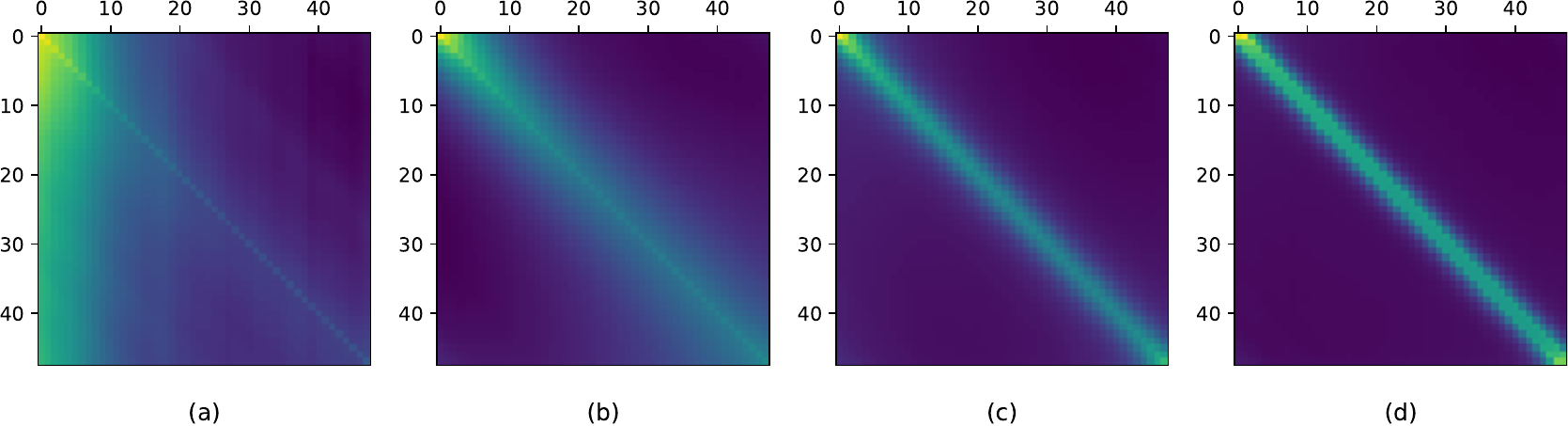}}
        \caption{Attention matrices of single layer vanilla Transformer. (a), (b), (c), (d) indicate the attention matrices when trained using 1\%, 10\%, 50\% 100\% of the PhysioNet data set.}
        \label{figure:1}
    \end{figure*}

The main contribution of this paper is that from the idea of underlying monotonic temporal bias in time series, we have designed the SAT-Transformer that can learn to exploit these regularities directly. In section \ref{method}, we show some direct supporting evidence for our initial hypothesis in addition to performance gains in section \ref{experiments}. In addition, we show multiple ablation experiments and extensions of our model to gain more insight into the performance of our model in terms of the temporal regularities of the data sets. Through our experiments, we also show that our model can be more useful, especially in situations with a limited amount of data. We believe that the results of our paper suggest a direction to explore, which is, rather than expecting neural network models to learn from scratch, exploiting temporal biases of time series can result in room for improvement. 

\section{Related Works}
\label{relwrks}
    \subsection{Structural Bias of Model and Generalization}
    Structural bias, including inductive bias of machine learning models, has been studied ever since the field of machine learning was studied. Researchers have gradually realized that for different structures of data, different kinds of structural bias must be applied to models to achieve a jump in performance. With this wisdom, researchers have developed models such as CNNs \cite{10.5555/646469.691875}, RNNs \cite{osti_6910294}, Graph Neural Networks \cite{1555942} or Transformers \cite{vaswani2017attention} for better prediction performance on data sets such as images, time series, and natural language. 
    
    More recently, with very similar philosophies, enabling models or learning agents to generalize to a more sophisticated task or different data formats is being explored extensively. Specifically, with the right inductive bias, one can train a reinforcement learning agent to glue blocks together to build a tower \cite{hamrick2018relational} or perform better in transfer learning problems \cite{li2018explicit}. In some tasks, the generalization of trained models differs not only in performance but also qualitatively \cite{McCoy2018}. These works altogether suggest that for specific types of tasks, specific structural or inductive biases are required.
    
    However, more evidences supporting that these structural biases might not be absolutely necessary are being discovered. A recent work done by \cite{dosovitskiy2021image} shows that Transformers can be trained to outperform the state-of-the-art CNN models on image data if the training set is extremely large. From these experiments, an intuitive conclusion can be drawn about the relation between the size of the data set and the flexibility of the model. That is, a more flexible model can eventually perform better than highly structural models when enough data are given.

    \subsection{Event Prediction in Medicine Using EHR}
    It is widely known in medicine that prediction or detection of deterioration in the ICU can contribute to better patient management, which leads to better outcomes \cite{power2014try}. While this is a critical issue, it is still a challenging task. Various machine learning-based approaches have been suggested recently to tackle these issues and discover the underlying characteristics of those patients. 
    
    Most of the works concentrate on improving the prediction performance by better imputing missing values, which have always been considered the biggest problem in EHR data analysis. The GRU-D, one of the earliest and most prominent works, predicts missing values based on the tendency that missing variables become closer to specific values as time goes by \cite{che2018recurrent}. This is done by exponential decay given to missing values and hidden states to specific values. The GRU-D outperforms vanilla GRU on the benchmark data sets by a significant margin. Phase LSTM introduces an additional time gate inside the LSTM cell. The time gate regularizes the access to the hidden state, enabling the hidden state to preserve the historical information for a more extended period \cite{neil2016phased}. In another work by \cite{li2016scalable}, the missing values were estimated using the Gaussian Process. The Gaussian Process and prediction network are trained together, where the output of the Gaussian Process is used as the input of the RNN. An extension of this idea is proposed in \cite{futoma2017learning}, which applies a multivariate Gaussian Process for imputation. Recently, \cite{shukla2019interpolation} proposed an interpolation prediction network that applies multiple semi-parametric interpolation processes to obtain regularly sampled time series data. SeFT proposed by \cite{horn2020set} attempts to solve this problem from a different perspective. It directly predicts the events without the imputation process by relaxing the sequence order condition of the input data.

    As the Transformers are becoming more popular in natural language processing and computer vision, many recent works utilize the Transformers to solve medically critical tasks using EHR data. In \cite{TransformEHR}, an encoder-decoder framework with a novel pretraining is proposed to enhance the detection of pancreatic cancer and self-harm in patients with post-traumatic stress disorder. Similarly, \cite{Meng_2021} utilized bidirectional representation learning with the Transformer architecture to predict depression from EHR data. In addition, many interesting works attempt to overcome the structural limitations of the Transformer architecture. In \cite{li2021hibehrthierarchicaltransformerbasedmodel}, the authors showed that applying a hierarchical Transformer on EHR data to expand the receptive field of the model better incorporates long-term dependencies. Because EHR data is highly irregular in time intervals, applying the Transformer directly can result in some suboptimal performance. To mitigate these issues, \cite{peng2021sequentialdiagnosispredictiontransformer} employs the Neural Ordinary Differential Equation in conjunction with the Transformer architecture, and \cite{tipirneni2022selfsupervisedtransformersparseirregularly} invented a novel Continuous Value Embedding technique to embed continuous time variables instead of applying discretization methods.

\section{Proposed Method}
\label{method}

    \subsection{Learning Underlying Structure of a Time Series}
    \label{reveal}
    We hypothesized that time series data have an underlying structure regardless of the prediction tasks. To elaborate, for a time series $\{X_t\}_{t=1}^{T}$ {\it interrelation} between time stamps $i$ and $j$ are larger for closer time stamps. Note that we do not mean this in a mathematically rigorous manner. Although this hypothesis seems rather trivial, learning relevant structures without inductive bias requires a tremendous amount of data \cite{dosovitskiy2021image}. In order to observe this behavior experimentally, we trained an extremely simple single-block transformer with a minimal amount of parameters. Though the model is too small to be useful for any real-world tasks, it is sufficient for the purpose of testing our hypothesis considering the size of the available data sets. The target task was guessing the value of a randomly chosen position on the eICU data set \cite{devlin2018bert}, which is explained in detail in section \ref{datasets}. As shown in Figure \ref{figure:1}, we observed how the values of the attention matrix change with respect to the percentage of the data set used in training. The values along the diagonal of the attention matrix can be represented as the {\it interrelation} of a value at a time step to itself. We observed that the attention values converged to the diagonal as larger data samples were used. Therefore, from this experiment, we drew the following conclusions. First, as we proposed in our original hypothesis, there is indeed a higher dependency on time stamps that are closer. Second, learning this intricate regularity requires quite a large amount of data, even for this extremely simple model. 
    
    Inspired by our original hypothesis and these elementary experiments, we devised temporal kernels for the Transformers' attention matrices. Although the experiments seem to suggest only monotonic kernels, we believe there can be additional unobserved dependencies. For this reason, the kernels should be designed with additional domain knowledge.
    
    \subsection{Architecture of the Attention Matrix}
    \label{attention}
    We first considered kernels taking the form of exponential covariance function:
    
     \begin{equation}
         \mathcal{C}_e(X_{t_1}, X_{t_2}) = \sigma^2 e^{-{(\alpha h)}^\beta}
         \label{equation:1}
     \end{equation}
    
    where $h = |t_1 - t_2|$ for time stamps at $X_{t_1}$ and $X_{t_2}$ and $\alpha$ and $\beta$ are learnable parameters. Additionally, in this paper, we considered periodic kernels because the medical data sets used in our main experiments are highly periodic: 
    
    \begin{equation}
        \mathcal{C}_p(X_{t_1}, X_{t_2}) = \sigma^2 e^{-2 \alpha^2 \sin^2(\frac{\pi h}{\beta})}
        \label{equation:2}
    \end{equation}

    where $h = |t_1 - t_2|$ for time stamps at $X_{t_1}$ and $X_{t_2}$ and $\alpha$ and $\beta$ are learnable parameters as equation \ref{equation:1}. Note that in equation \ref{equation:1} and equation \ref{equation:2}, though the names of learnable parameters are the same, this is just due to the simplicity of the notation, and thus, they have different values in actual implementation. Also, we first normalize the data using z-normalization, resulting in $\sigma^2 = 1$ for simplicity of notation throughout this paper. 
    
    Let $\mathbf{A} \in \mathbb{R}^{T \times T}$ be the attention matrix with $\mathbf{V} = \{v_i\} \in \mathbb{R}^{T \times d_i}$ as the input sequence and $\mathbf{W_1}, \mathbf{W_2} \in \mathbb{R}^{d_i \times d_k}$ be the learned matrices for transforming $\mathbf{V}$ into query and key representations of $\mathbf{Q} = \mathbf{VW_1}$, $\mathbf{K} = \mathbf{VW_2}$. The kernels defined in equations \ref{equation:1} and \ref{equation:2} are applied to each element of the query and key matrices as follows:
    
    % \begin{equation}
    %     \mathcal{K}(X_{t_1}, X_{t_2}) = \mathcal{K}_e(X_{t_1}, X_{t_2})\mathcal{K}_p(X_{t_1}, X_{t_2})
    % \end{equation}.

    \begin{equation}
        \mathbf{\widehat{Q}}_{i, j} = \mathcal{C}_e(\mathbf{Q}_{i, i}, \mathbf{Q}_{i, j})   \mathbf{Q}_{i, j}
        \label{equation:3}
    \end{equation}
    
    \begin{equation}
        \mathbf{\widehat{K}}_{i, j} = \mathcal{C}_p(\mathbf{K}_{i, i}, \mathbf{K}_{i, j}) \mathbf{K}_{i, j}
        \label{equation:4}
    \end{equation}
    
    getting the attention matrix with kernelization as follows:
    
    \begin{equation}
        \mathbf{\widehat{A}} = \textit{softmax} \left ( \frac{\mathbf{\widehat{Q}}\mathbf{\widehat{K}}^\top}{\sqrt{d_k}} \right)
        \label{equation:5}
    \end{equation}
    
    However, applying these kernels element-wise is computationally inefficient because it harms parallelism. We have stacked kernel elements in a matrix format to solve this issue by applying vectorized multiplication. Since kernels are not dependent on the input's values but rather on their time stamps, a fixed kernel matrix can be derived for an attention matrix of fixed size. Therefore, the kernel matrices $\mathbf{C}^{(e)} \in \mathbb{R}^{T \times T}$ and $\mathbf{C}^{(p)} \in \mathbb{R}^{T \times T}$ can be defined as following:
    
    \begin{equation}
        \mathbf{C}_{i, j}^{(e)} = e^{-{(\alpha |i-j|)}^\beta} 
        \label{equation:6}
    \end{equation}
    
    \begin{equation}
        \mathbf{C}_{i, j}^{(p)} = e^{-2 \alpha^2 \sin^2(\frac{\pi |i-j|}{\beta})} 
        \label{equation:7}
    \end{equation}
    
    With these matrix versions of kernels, we can rewrite the attention matrix to take advantage of vectorization. 
    
   \begin{equation}
        \mathbf{\widehat{A}} = \textit{softmax} \left ( \frac{(\mathbf{C}^{(e)}\odot\mathbf{Q})(\mathbf{C}^{(p)}\odot\mathbf{K})^\top}{\sqrt{d_k}} \right )
        \label{equation:8}
    \end{equation}

    Where $\odot$ denotes element-wise multiplication. 

\begin{table*}[h!]
    \centering
        \resizebox{17.2cm}{!}{
            \begin{tabular}{l c c | c c | c c | c c | c c | c}
                \toprule
                & \multicolumn{2}{c}{PhysioNet} & \multicolumn{2}{c}{MIMIC-III} & \multicolumn{2}{c}{eICU-HF} & \multicolumn{2}{c}{eICU-RF} & \multicolumn{2}{c|}{eICU-KF} & \multirow{2}{*}{ms/iter} \\
                \cline{2-11}
                & AUPRC & AUROC  & AUPRC  & AUROC & AUPRC & AUROC & AUPRC & AUROC & AUPRC & AUROC \\
                \hline
                GRU-S & 15.0 ± 0.5 & 82.5 ± 0.4  & 51.6 ± 0.6 & 85.4 ± 0.4& 2.46 ± 0.11  & 77.0 ± 0.5& 3.48 ± 0.01  & 71.8 ± 0.1& 2.65 ± 0.01  & 76.8 ± 0.5  & 24.8\\
                GRU-D & 14.9 ± 0.6 & 82.8 ± 0.3  & 52.8 ± 0.5 & 86.1 ± 0.7& 2.92 ± 0.05  & 80.1 ± 0.2& 3.42 ± 0.13  & 72.3 ± 0.7& 2.60 ± 0.16 & 78.3 ± 0.7  & 41.9 \\
                IP-Nets	& 15.3 ± 0.2 & 82.7 ± 0.1 & 51.8 ± 0.9 & 85.4 ± 0.1 & 2.92 ± 0.01 & 80.8 ± 0.1  & 3.30 ± 0.02& 73.4 ± 0.1 & 2.31 ± 0.03 & 77.1 ± 0.1   & 25.8  \\
                Transformer & 15.0 ± 0.2 & 81.4 ± 0.2 & 49.7 ± 0.3 & 84.8 ± 0.2 & 2.92 ± 0.08 & \textbf{81.3 ± 0.1} & 3.58 ± 0.06 & 73.5 ± 0.2  & 3.09 ± 0.28 & 80.8 ± 0.3  & 6.71  \\
                SeFT & 13.3 ± 0.2 & 82.0 ± 0.1 & 46.2 ± 0.1 & 85.1 ± 0.3  & \textbf{3.16 ± 0.01} & 78.3 ± 0.7 & 3.32 ± 0.03 & 72.4 ± 0.3 & 3.08 ± 0.30 & 78.3 ± 0.1  & 3.78 \\
                SAT-Trans & \textbf{16.7 ± 0.3}  & \textbf{83.0 ± 0.7}  & \textbf{53.7 ± 0.5} & \textbf{86.4 ± 0.3}  & 3.10 ± 0.09 & 81.0 ± 0.3   & \textbf{3.80 ± 0.03} & \textbf{73.7 ± 0.2} & \textbf{3.33 ± 0.03} & \textbf{81.0 ± 0.5}  & 7.23 \\
                \toprule
            \end{tabular}
        }
        \caption{Performance of models for each data set and task.}
    \label{table:1}
\end{table*}

\section{Experiments}
\label{experiments}
    
    \subsection{Data Sets}
    \label{datasets}
    We used three different open EHR data sets, which contain multiple labels described below.
    
    The PhysioNet Challenge 2019 data set \cite{reyna_josef_jeter_shashikumar_moody_westover_sharma_nemati_clifford_2019,reyna_josef_jeter_shashikumar_westover_nemati_clifford_sharma_2020, PhysioNet} consist of hourly clinical variables collected from intensive care unit (ICU) of two hospital systems. Clinical variables include 8 vital signs, 26 laboratory values, and 6 demographic information. The data set contains 40,336 patients with the number of rows of 1,424,171. With this data set, the task is to predict sepsis within 12 hours, where the onset of sepsis is defined by sepsis-3 criteria \cite{reyna_josef_jeter_shashikumar_moody_westover_sharma_nemati_clifford_2019,singer2016third}. Since the original data set does not contain event labels in the time window of 6 hours to 12 hours prior to the onset of sepsis, we inserted additional labels to the corresponding points. The number of septic patients is 2,932, accounting for 7.2\% of entire subjects.
    
    MIMIC-III \cite{johnson2016mimic} is a multivariate clinical time series database collected at Beth Israel Deaconess Medical Center. We processed the data set for mortality prediction based on the method defined at \cite{harutyunyan2019multitask}. The extracted data set contains 21,139 ICU stays, and each ICU stay includes 17 clinical variables with measurement intervals of 1 hour. The objective is to predict in-hospital mortality using the measurements from the first 48 hours of ICU stay. The number of positive cases is 2,797, which is 13.2\% of the total number of patients.
    
    The eICU Collaborative Research Database \cite{pollard2018eicu} is a freely available multi-center database containing over 200,000 admissions to ICU. The database was collected from 335 units at 208 hospitals in the US. We processed the data set following the procedure defined at the PhysioNet challenge 2019 \cite{reyna_josef_jeter_shashikumar_moody_westover_sharma_nemati_clifford_2019}. Additionally, we excluded patients if there was at least one unmeasured time stamp. Among 200,859 admissions, 132,112 admissions with 7,692,965 rows are extracted as a result. Each of the data points includes 40 variables measured in one-hour intervals. Here, we defined multiple tasks of predicting Heart failure, Respiratory failure, and Kidney failure within 12 hours prior to diagnosis. Each event is set based on the ICD-9 code (HF:428.0, RF:518.81, KF:584.9) on EHR. The number of patients is 4,286 (3.24\%), 8,279 (6.27\%), and 4,357 (3.30\%) for each task.
   
    \subsection{Models and Settings}
    We compared our method to the widely used, best-performing models in time series prediction.
    (1) GRU-Simple: an extension of GRU that takes time series and information about missing variables as input.
    (2) GRU-D: an extension of GRU that implements hidden state decays and missing value decays, encouraging convergence to the mean value. 
    (3) Interpolation Network: predicts missing values using adjacent measurement and radial basis functions. 
    (4) SeFT: encodes each observation separately as an unordered set and pooling them together in the attention layer. 
    (5) Transformer: uses self-attention matrices, which considers the entire sequence to encode the relationship between each time stamp. 
    
    Unlike the MIMIC-III data set, the PhysioNet and eICU data set does not contain a fixed number of sequence lengths, making the prediction to be made using varying lengths of time series. To limit the memory requirement, we used the window of 48 hours prior to prediction points. For all data sets, we used 80\% of the instance as a training set and the rest as a testing set. Within the 80\% of the training set, 20\% of the data are randomly selected as the validation set. For PhysioNet and MIMIC-III, the test sets were constructed to be exactly the same as \cite{horn2020set} for a fair comparison. The test set for eICU was created by randomly selecting 20\% of the data samples. 
    
    Since EHR data sets are highly skewed, we constructed a balanced batch by sampling event and non-event cases to the same ratio during training. The area under the precision-recall curve (AUPRC) was used to measure the validation performance of the model to determine the best hyperparameters. We stopped training if AUPRC does not improve for 30 epochs. In the tuning process, more than 200 hyperparameter sets are explored for each model trained on PhysioNet and MIMIC-III data sets. In addition, we repeated the experiments 3 times using different random seeds for stable performance. The eICU database contains more than 7,000,000 rows, making it difficult to conduct extensive hyperparameter searches. Thus, we explored smaller hyperparameter space where each hyperparameter set was tested twice with different random seeds. Training was stopped if the increase of AUPRC was not observed for 5 epochs. We tested around 50 hyperparameter sets for each model for each task from the eICU data set. Further, all ablation studies and extension experiments are done on the PhysioNet data set. As a result, the baselines shown throughout this paper are at least as good as the baselines used in many works in literature, if not better. All the details discussed in this section, including baseline references and search space for hyperparameters, are included in the provided supplementary.
    
    \subsection{Results}
    Table \ref{table:1} shows the performance of the SAT-Transformer against other well-known models with the best performance up to date. Regarding AUPRC, which is the most appropriate metric for unbalanced data sets, SAT-Transformer performs significantly better than other models in almost every task across multiple data sets. In terms of AUROC, while it is highly affected by the imbalance in the data sets, SAT-Transformer still performs best in most of the tasks. Additionally, while the second best-performing model significantly varies for different tasks and data sets, SAT-Transformer achieves that best performance significantly across all tasks. To measure computational efficiency, time (in milliseconds) per iteration has been measured and averaged over multiple trials. As a result, the SAT-Transformer is comparable to or only slightly slower than the vanilla Transformer and much more efficient than other RNN-based models. 
    
\begin{figure}[h!]
      \centering
      \centerline{\includegraphics[width=8.55cm]{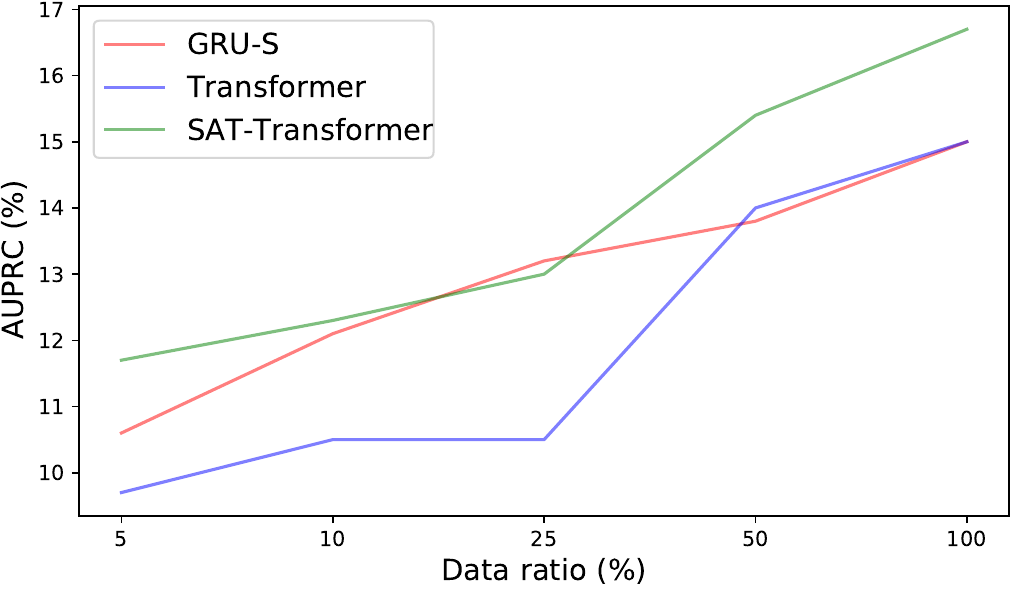}}
        \caption{Performance of each model with respect to reduction of data set size.}
        \label{figure:2}
    \end{figure}

    To demonstrate the effect of the encoded inductive bias of the SAT-Transformer according to the size of the data set, we examined the drop in performance when only a subset of data is used in training. We selected GRU-Simple, vanilla Transformer as a representative in each category. As shown in Figure \ref{figure:2}, the performance difference between SAT-Transformer and GRU-Simple becomes significant as the size of the data set increases. Another point to note is that the performance increase of the vanilla Transformer becomes more significant as more data samples are used, achieving similar performance as GRU-Simple when 100\% of the data set was used. A recent study in \cite{dosovitskiy2021image} suggests that models with more flexibility or less inductive bias can eventually outperform models with more structures in the case of extremely large data sets. Similar behavior can be observed when the two models are compared with SAT-Transformer. Since GRU-Simple is highly structured due to the recurrent architecture when dealing with time series and Transformers are more flexible, a less dramatic drop in the performance of GRU-Simple makes sense. The SAT-Transformer, however, can be thought of as being almost as flexible as the Transformer while it is capable of maintaining the inductive bias of the RNNs. When given an extremely large data set, the vanilla Transformer may eventually outperform every model, but the performance difference shown in Table \ref{table:1} suggests that this is currently almost infeasible.
    
    % \begin{table*}[h!]
    %     \centering
    %         \resizebox{17.2cm}{!}{
    %             \setcounter{table}{3}
    %             \begin{tabular}{l c c | c c | c c | c c | c c}
    %                 \toprule
    %                 & \multicolumn{2}{c}{PhysioNet} & \multicolumn{2}{c}{MIMIC-III} & \multicolumn{2}{c}{eICU-HF} & \multicolumn{2}{c}{eICU-RF} & \multicolumn{2}{c}{eICU-KF} \\
    %                 \cline{2-11}
    %                 & AUPRC& AUROC  & AUPRC & AUROC & AUPRC& AUROC  & AUPRC & AUROC & AUPRC& AUROC \\
    %                 \hline
    %                 Non-Adaptive &  \textbf{16.7 ± 0.3}  & 83.0 ± 0.7  & \textbf{53.7 ± 0.5} & \textbf{86.4 ± 0.3}  & \textbf{3.10 ± 0.09} & \textbf{81.0 ± 0.3}   & \textbf{3.80 ± 0.03} & \textbf{73.7 ± 0.2} & \textbf{3.33 ± 0.03} & \textbf{81.0 ± 0.5} \\

    %                 Adaptive &  16.4 ± 0.3 & \textbf{83.3 ± 0.3} & 53.2 ± 0.6 & 86.0 ± 0.3 & 3.06 ± 0.06 & \textbf{81.0 ± 0.4} & 3.70 ± 0.01 & 72.8 ± 0.2 & 3.26 ± 0.01 & 80.7 ± 0.1 \\
    %                 \toprule
    %             \end{tabular}
    %         }
    %         \caption{Effect of temporal feature adaptive kernels on performance}
    %     \label{table:4}
    % \end{table*}
    
    \begin{figure*}[h!]
        \centering
        \centerline{\includegraphics[width=17cm]{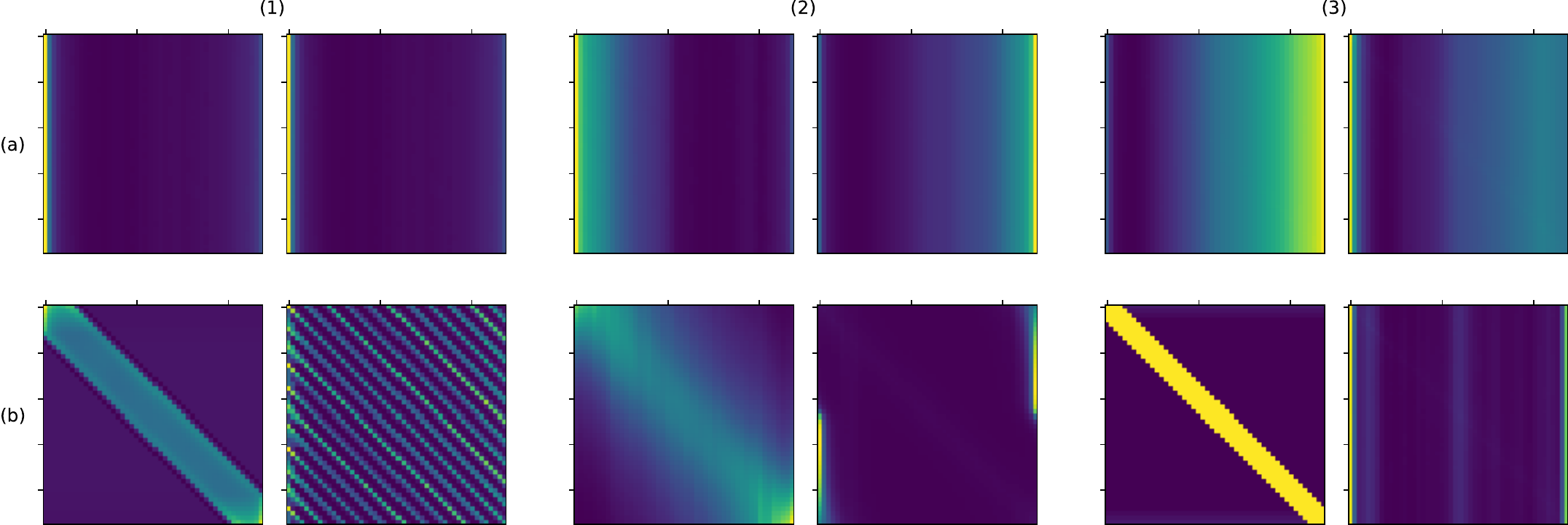}}
        \caption{Attention matrices from SAT-Transformer and vanilla Transformer for all three layers in two different heads. (1), (2), and (3) denote layer 1, layer 2, and layer 3 each. (a) denotes the vanilla Transformer, and (b) denotes the SAT-Transformer. The Left and right figures in each group indicate two different heads.}
        \label{figure:3}
    \end{figure*}
    
    \subsection{Ablation Study and Extensions}
    The periodic kernels have been added due to the prior domain knowledge of our medical data sets. To see the effect of this, we performed an ablation study using only exponential kernels and only periodic kernels. The experiments are done on PhysioNet data sets with sepsis prediction as the target task. As the results summarized in Table \ref{table:2} show, the model performs better than the vanilla Transformer when any kernels are used. Comparing exponential kernels and periodic kernels, the performance gain resulting from exponential kernels is more significant than periodic kernels.
    
    \begin{table}[h!]
        \centering
            \resizebox{6.5cm}{!}{
            \setcounter{table}{1}
                \begin{tabular}{l c c}
                    \toprule
                    Kernels & AUPRC & AUROC  \\
                    \hline
                    No Kernel  &  15.0 ± 0.2 & 81.4 ± 0.2 \\
                    Exp & 16.3 ± 0.6 & \textbf{83.2 ± 0.2}  \\
                    Periodic& 15.9 ± 0.3 & 81.7 ± 0.1   \\
                    Exp \& Periodic & \textbf{16.7 ± 0.3} &  83.0 ± 0.7  \\
                    \toprule
                \end{tabular}
            }
            \caption{Performance of SAT-Transformer according to the use of different kernels.}
        \label{table:2}
    \end{table}

    We explored two types of additional extensions that can be applied to SAT-Transformer. First, to demonstrate the compatibility of SAT-Transformer as a module with the existing framework, we added the Interpolation Network module defined in section 3.2.1 of \cite{shukla2019interpolationprediction} on top of our model. We also trained a vanilla transformer with an Interpolation Network added to the top for comparison. These experiments are done on PhysioNet data sets with sepsis prediction as the target task. As shown in Table \ref{table:3}, both models with Interpolation Network gain additional performance as expected. 
    
    \begin{table}[h!]
        \centering
            \resizebox{7.5cm}{!}{
                \setcounter{table}{2}
                \begin{tabular}{l c c}
                    \toprule
                    Models & AUPRC & AUROC \\
                    \hline
                    Transformer &  15.0 ± 0.2  & 81.4 ± 0.2 \\
                    Transformer \& IP-Net &  15.4 ± 0.6  & 82.0 ± 0.3\\
                    SAT-Transformer &  16.7 ± 0.3 & 83.0 ± 0.7 \\
                    SAT-Trans \& IP-Net  & \textbf{17.5 ± 0.2} & \textbf{83.7 ± 0.3} \\
                    \toprule
                \end{tabular}
            }
            \caption{Performance of Transformer models with Interpolation Network added on top.}
        \label{table:3}
    \end{table}

    Second, because time series data can be highly diverse, kernels that depend on temporal features can be beneficial. For example, in certain prediction tasks involving EHR data, the length of the vital sequence can be a significant factor affecting the prediction performance \cite{li2019tasp}. Therefore, we designed temporal feature adaptive kernels that the learnable parameters $\alpha$ and $\beta$ from equation \ref{equation:6} and \ref{equation:7} can be adaptive. Specifically, $\alpha$ and $\beta$ are computed from a linear model taking temporal features for each data point $\{v_t\}_{t=1}^T$ as an input. The temporal feature inputs are defined as vectors of mean, standard deviation, sequence length, and average time interval $(\frac{t_T - t_1}{T})$. The experiments are done on every EHR data set as presented in Table \ref{table:1}, and the results are presented in Table \ref{table:4}. There are some minor improvements for some tasks using adaptive kernels, but they are relatively inconsistent.  

    \begin{table*}[h!]
        \centering
            \resizebox{17.2cm}{!}{
                \begin{tabular}{l c c | c c | c c | c c | c c}
                    \toprule
                    & \multicolumn{2}{c}{PhysioNet} & \multicolumn{2}{c}{MIMIC-III} & \multicolumn{2}{c}{eICU-HF} & \multicolumn{2}{c}{eICU-RF} & \multicolumn{2}{c}{eICU-KF} \\
                    \cline{2-11}
                    & AUPRC& AUROC  & AUPRC & AUROC & AUPRC& AUROC  & AUPRC & AUROC & AUPRC& AUROC \\
                    \hline
                    Non-Adaptive &  \textbf{16.7 ± 0.3}  & 83.0 ± 0.7  & \textbf{53.7 ± 0.5} & \textbf{86.4 ± 0.3}  & \textbf{3.10 ± 0.09} & \textbf{81.0 ± 0.3}   & \textbf{3.80 ± 0.03} & \textbf{73.7 ± 0.2} & \textbf{3.33 ± 0.03} & \textbf{81.0 ± 0.5} \\

                    Adaptive &  16.4 ± 0.3 & \textbf{83.3 ± 0.3} & 53.2 ± 0.6 & 86.0 ± 0.3 & 3.06 ± 0.06 & \textbf{81.0 ± 0.4} & 3.70 ± 0.01 & 72.8 ± 0.2 & 3.26 ± 0.01 & 80.7 ± 0.1 \\
                    \toprule
                \end{tabular}
            }
            \caption{Effect of temporal feature adaptive kernels on performance}
        \label{table:4}
    \end{table*}

\section{Discussion}
\label{discussion}

From our original hypothesis, we have proposed the SAT-Transformer, verifying that it significantly outperforms other best-performing models up to date in multiple prediction tasks across various medical data sets. In this section, we further examine the learned attention matrices and kernels of SAT-Transformer from the perspective of the original hypothesis. 

We first examine the learned attention matrices of the SAT-Transformer and compare them against the attention matrices of the vanilla Transformer. To show the effect across every element of the model, we present the attention matrices for all three layers for two different heads. As shown in Figure \ref{figure:3}, many attention matrices in SAT-Transformer reveal higher \textit{interrelation} between closer time stamps. In addition, the effects of periodic kernels are observed in some attention matrices. In contrast, in the exact same matrices in the vanilla transformer, such behaviors are unobserved, showing rather similar attention maps across multiple heads and layers. From these observations, we conclude that with the assistance of the temporal kernels, the model was able to learn a more relevant attention scheme for given medical data sets and tasks of interest. In addition, with the previous experiment shown in Figure \ref{figure:1} and attention matrices shown in \ref{figure:3}, we demonstrate supporting pieces of evidence for our original hypotheses. 

\begin{figure}[h!]
    \centering
    \centerline{\includegraphics[width=8.2cm]{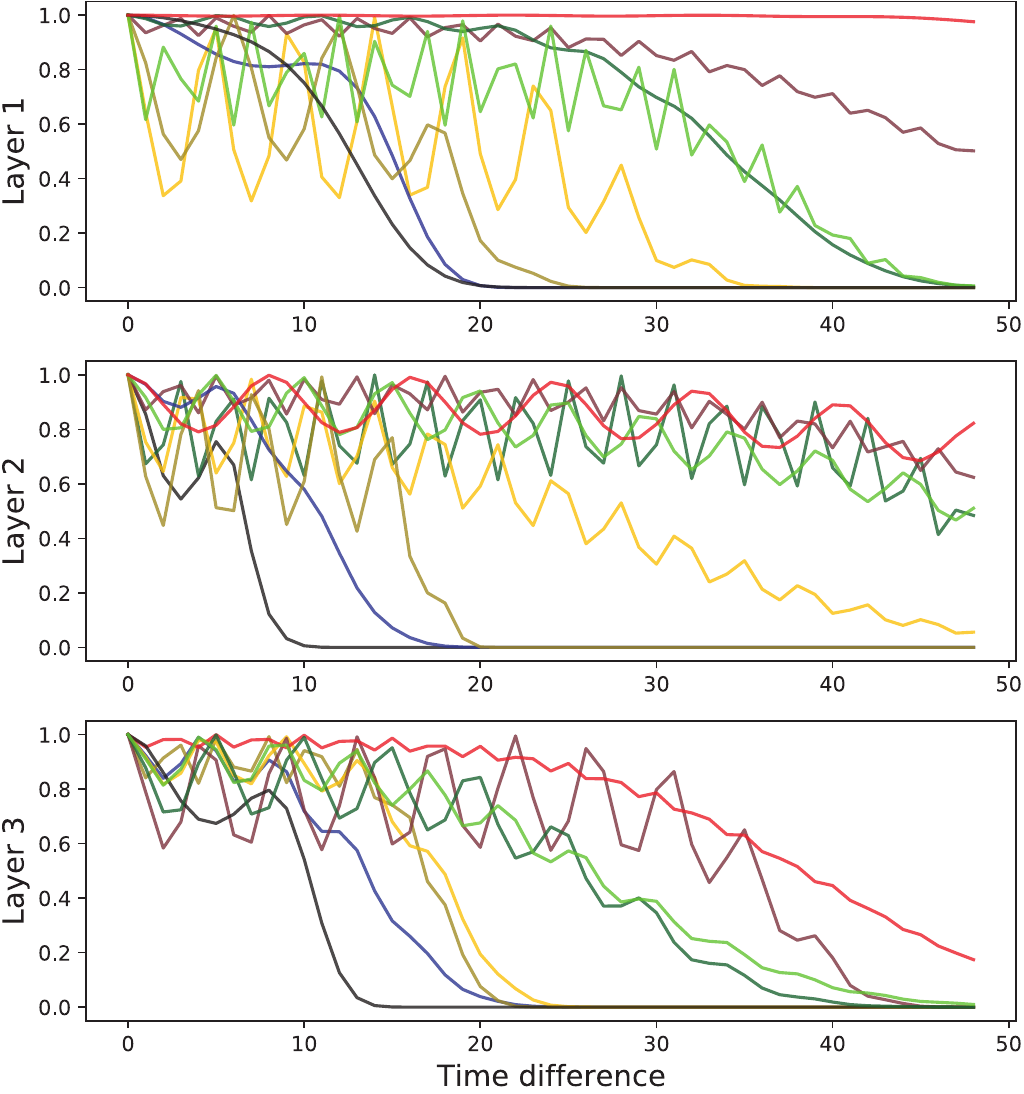}}
    \caption{Representative behavior of learned kernels for all three layers of SAT-Transformer. The different colors represent the behaviors of the kernels of different attention heads.}
    \label{figure:4}
\end{figure}

In Figure \ref{figure:4}, we show the shape of learned kernels without the effect of the attention for multiple heads in all three layers. Here, we observed that diverse kernels are learned for different attention heads. In many cases, the effect of exponential and periodic kernels coexist, but one of the kernels is dropped in some cases. Kernels in some heads are tuned to have almost no impact by obtaining values that are mostly close to $1$ when necessary.

A recent study suggested that attention matrices in Transformers with multiple heads tend to become very similar to each other, collapsing into a single matrix \cite{voita2019analyzing}. This work suggests that because of this lack of diversity, Transformers experience some limitations in its potential. In SAT-Transformer, due to the effect of kernels, we found that attention matrices became much more diverse compared to vanilla Transformer. Although this may not be the main reason SAT-Transformer reaches exceptional performance, we believe this might be one of the minor reasons.

\section{Conclusion}
\label{conclusion}

In this paper, we propose the SAT-Transformer and the elemental philosophy. We believe that our model as well as the analysis of the underlying phenomenon, provides an intuition and potential to be generalized to various time series data sets. Compared to other best-performing models known to date, SAT-Transformer has achieved outstanding gains in performance. In addition, we made an important observation of the fact that time series data arising in nature have \textit{interrelational} bias between close data points. By taking advantage of these underlying structures, we were able to gain a significant amount of performance. Finally, although many evidences, including ours, suggest that, these regularities can eventually be learned with enough data. However, this often requires an extremely large amount of data, which is still infeasible. In this respect, imposing the right kind of inductive bias, such as the one in SAT-Transformer, can be an extremely powerful concept in the current era. In future work, we would like to apply variations of our model to multiple time series data sets, exploring more efficient and sophisticated methods to construct kernels with temporal priors.

\section*{Acknowledgments}
This work was supported by the Korea Medical Device Development Fund grant funded by the Korean government (the Ministry of Science and ICT, the Ministry of Trade, Industry and Energy, the Ministry of Health \& Welfare, the Ministry of Food and Drug Safety) (Project Number: 1711137961, RS-2020-KD000030).


\bibliography{sat_transformer}
\bibliographystyle{icml2021}

\clearpage
\newpage

\renewcommand{\thetable}{\Alph{table}}
\renewcommand{\thesection}{\Alph{section}}

\setcounter{section}{0}
\setcounter{table}{0}
\section{Appendix}
\subsection{Training and Hyperparameter Search}

All of our implementations are in Tensorflow 1.13 and all experiments were done in NVIDIA Geforce RTX 2080Ti. For each model, we provide the following details on the explored hyperparameter search space. In order for more extensive hyperparameter tuning, we first performed smaller sets of test experiments to define the ranges of the values to search from. We then performed a grid search over all hyperparameter combinations from the found ranges, which are provided in the lists below. Because the eICU data set is much larger than other data sets, we defined a smaller search space due to the limitations in computational resources. The selected hyperparameters are presented in the Table \ref{table:1}.  

The SAT-Transformer must learn kernel parameters in addition to the regular parameters of the Transformer. For ease of hyperparameter search, we optimized the Transformer part and the kernel part separately. The learning rate2 denotes the learning rate used to optimize kernel parameters which is given as multiplication factor to the learning rate used to optimize Transformer part of the network.

\textbf{GRU-Simple}

1) Physionet/MIMIC-III
\begin{flalign*}
    &\quad \text{learning rate} \in \{0.0001,\ 0.0002,\ 0.0005\} &\\
    &\quad \text{batch size} \in \{32,\ 64,\ 128\} &\\
    &\quad \text{hidden units} \in \{128,\ 256,\ 512\} &\\
    &\quad \text{dropout} \in \{0,\ 0.1,\ 0.2\} &\\
    &\quad \text{dropout recurrent} \in \{0,\ 0.1,\ 0.2\}
\end{flalign*}

2) eICU
\begin{flalign*}
    &\quad \text{learning rate} \in \{0.0001,\ 0.0002,\ 0.0005 \} &\\
    &\quad \text{batch size} \in \{32,\ 64\} &\\
    &\quad \text{hidden units} \in \{256,\ 512\} &\\
    &\quad \text{dropout} \in \{0,\ 0.1\} &\\
    &\quad \text{dropout recurrent} \in \{0,\ 0.1\}
\end{flalign*}

\textbf{GRU-D}

1) Physionet/MIMIC-III
\begin{flalign*}
    &\quad \text{learning rate} \in \{0.0001,\ 0.0002,\ 0.0005 \} &\\
    &\quad \text{batch size} \in \{32,\ 64,\ 128\} &\\
    &\quad \text{hidden units} \in \{128,\ 256,\ 512\} &\\
    &\quad \text{dropout} \in \{0,\ 0.1,\ 0.2\} &\\
    &\quad \text{dropout recurrent} \in \{0,\ 0.1,\ 0.2\}
\end{flalign*}

2) eICU
\begin{flalign*}
    &\quad \text{learning rate} \in \{0.0001,\ 0.0002,\ 0.0005 \} &\\
    &\quad \text{batch size} \in \{32,\ 64\} &\\
    &\quad \text{hidden units} \in \{256,\ 512\} &\\
    &\quad \text{dropout} \in \{0,\ 0.1\} &\\
    &\quad \text{dropout recurrent} \in \{0,\ 0.1\}
\end{flalign*}

\textbf{Interpolation Networks}

1) Physionet/MIMIC-III
\begin{flalign*}
     &\quad \text{learning rate} \in \{0.0002,\ 0.0005\} &\\
     &\quad \text{batch size} \in \{16,\ 32,\ 64 \} &\\
     &\quad \text{hidden units} \in \{128,\ 256,\ 512\} &\\
     &\quad \text{dropout} \in \{0,\ 0.1,\ 0.2\} &\\
     &\quad \text{dropout recurrent} \in \{0,\ 0.1 \} &\\
     &\quad \text{reconstruction fraction} \in \{0.1,\ 0.2,\ 0.5\}
\end{flalign*}

2) eICU
\begin{flalign*}
    &\quad \text{learning rate} \in \{0.0002,\ 0.0005 \} &\\
    &\quad \text{batch size} \in \{32,\ 64\} &\\
    &\quad \text{hidden units} \in \{128,\ 256\} &\\
    &\quad \text{dropout} \in \{0.1,\ 0.2\} &\\
    &\quad \text{dropout recurrent} \in \{0.0,\ 0.1\} &\\
    &\quad \text{reconstruction fraction} \in \{0.2,\ 0.5\}
\end{flalign*}

\textbf{Transformer}

1) Physionet/MIMIC-III
\begin{flalign*}
    &\quad \text{learning rate} \in\{0.0002,\ 0.0005,\ 0,001\} &\\
    &\quad \text{batch size} \in \{32,\ 64\} &\\
    &\quad \text{num of layers} \in \{2,\ 3,\ 4\} &\\
    &\quad \text{num of heads} \in \{2,\ 4,\ 8\} &\\
    &\quad \text{hidden units} \in \{256,\ 512\} &\\
    &\quad \text{dropout} \in \{0,\ 0.1,\ 0.2\}
\end{flalign*}

2) eICU
\begin{flalign*}
    &\quad \text{learning rate} \in \{0.0005,\ 0.001\} &\\
    &\quad \text{batch size} \in \{32,\ 64\} &\\
    &\quad \text{num of layers} \in \{3\} &\\
    &\quad \text{num of heads} \in \{2,\ 4,\ 8\} &\\
    &\quad \text{hidden units} \in \{256,\ 512\} &\\
    &\quad \text{dropout} \in \{0.1,\ 0.2\}
\end{flalign*}

\textbf{SeFT}

1) Physionet/MIMIC-III
\begin{flalign*}
    &\quad \text{learning rate} \in \{0.0005,\ 0.001,\ 0.002\} &\\
    &\quad \text{batch size} \in \{128,\ 256\} &\\
    &\quad \phi \text{ (layers)} \in \{2,\ 3,\ 4 \} &\\
    &\quad \phi \text{ (width)} \in \{128,\ 256\} &\\
    &\quad \rho \text{ (layers)} \in \{2,\ 3,\ 4\} &\\
    &\quad \rho \text{ (width)} \in \{128,\ 256,\ 512\} &\\
    &\quad \psi \text{ (layer)}  \in \{2\} &\\
    &\quad \psi \text{ (width)}  \in \{64\} &\\
    &\quad \text{dot prod dim} \in \{128\} &\\
    &\quad \text{latent width: same as } \phi \text{ width} &\\
    &\quad \text{max time scale} \in \{1000\} &\\
    &\quad \text{pos dims} \in \{8\} &\\
    &\quad \text{dropout: randomly selected from } \{0.1,\ 0.2,\ 0.3,\ 0.4,\ 0.5\}
\end{flalign*}
    
2) eICU
\begin{flalign*}
    &\quad \text{learning rate} \in \{0.0005,\ 0.002\} &\\
    &\quad \text{batch size} \in \{128,\ 256\} &\\
    &\quad \phi \text{ (layers)} \in \{2,\ 4 \} &\\
    &\quad \phi \text{ (width)} \in \{128,\ 256\} &\\
    &\quad \rho \text{ (layers)} \in \{2,\ 3\} &\\
    &\quad \rho \text{ (width)} \in \{256,\ 512\} &\\
    &\quad \psi \text{ (layer)} \in \{2\} &\\
    &\quad \psi \text{ (width)} \in \{64\} &\\
    &\quad \text{dot prod dim} \in \{128\} &\\
    &\quad \text{latent width: same as } \phi \text{ width} &\\
    &\quad \text{max time scale} \in \{1000 \} &\\
    &\quad \text{pos dims} \in \{8 \} &\\
    &\quad \text{dropout: randomly selected from} \{0.1,\ 0.2,\ 0.3\}
\end{flalign*}

\textbf{SAT-Transformer}

1) Physionet/MIMIC-III
\begin{flalign*}
    &\quad \text{learning rate} \in \{0,0002,\ 0.0005,\ 0.001 \} &\\
    &\quad \text{batch size} \in \{32,\ 64\} &\\
    &\quad \text{num of layers} \in \{2,\ 3,\ 4\} &\\
    &\quad \text{num of heads} \in \{2,\ 4,\ 8\} &\\
    &\quad \text{hidden units} \in \{256,\ 512\} &\\
    &\quad \text{dropout} \in \{0,\ 0.1,\ 0.2\} &\\
    &\quad \text{learning rate2} \in \{\text{lrx20},\ \text{lrx50},\ \text{lrx100}\}&
\end{flalign*}

2) eICU
\begin{flalign*}
    &\quad \text{learning rate} \in \{0.0005,\ 0.001\} &\\
    &\quad \text{batch size} \in \{32,\ 64\} &\\
    &\quad \text{num of layers} \in \{3\} &\\
    &\quad \text{num of heads} \in \{2,\ 4,\ 8\} &\\
    &\quad \text{hidden units} \in \{256,\ 512\} &\\
    &\quad \text{dropout} \in \{0.1,\ 0.2\} &\\
    &\quad \text{learning rate2} \in \{\text{lrx20},\ \text{lrx50}\}& 
\end{flalign*}

\renewcommand{\cellalign}{l}
\begin{table*}[h!]
    \centering
        \resizebox{17.2cm}{!}{
            \begin{tabular}{l l | l | l | l | l}
                \toprule
                & \multicolumn{1}{c}{PhysioNet} & \multicolumn{1}{c}{MIMIC-III} & \multicolumn{1}{c}{eICU-HF} & \multicolumn{1}{c}{eICU-RF} & \multicolumn{1}{c}{eICU-KF} \\
                \hline
                GRU-Simple & \makecell{learning rate: 0.0002 \\ batch size: 32 \\ hidden units: 512 \\ dropout: 0.2 \\
                dropout recurrent: 0.0} & \makecell{learning rate: 0.0002 \\ batch size: 32 \\ hidden units: 256 \\ dropout: 0.1 \\ dropout recurrent: 0.0} & \makecell{learning rate: 0.0001 \\ batch size: 32 \\ hidden units: 512 \\ dropout: 0.1 \\ dropout recurrent: 0.1} & \makecell{learning rate: 0.0001 \\ batch size: 32 \\ hidden units: 256 \\ dropout: 0.1 \\ dropout recurrent: 0.1} & \makecell{learning rate: 0.0005 \\ batch size: 64 \\ hidden units: 256 \\ dropout: 0.1 \\ dropout recurrent: 0.1}\\ \hline
                
                GRU-D & 
                \makecell{learning rate: 0.0002 \\ batch size: 32 \\ hidden units: 512 \\ dropout: 0.2 \\ dropout recurrent: 0.2} & 
                \makecell{learning rate: 0.0002 \\ batch size: 32 \\ hidden units: 128 \\ dropout: 0.2 \\ dropout recurrent: 0.0} & 
                \makecell{learning rate: 0.0005 \\ batch size: 64 \\ hidden units: 512 \\ dropout: 0.1\\ dropout recurrent: 0.1} &
                \makecell{learning rate: 0.0005 \\ batch size: 32 \\ hidden units: 512 \\ dropout: 0.1 \\ dropout recurrent: 0.1} &
                \makecell{learning rate: 0.0005 \\ batch size: 64 \\ hidden units: 512 \\ dropout: 0.1 \\ dropout recurrent: 0.1}
                \\ \hline
                
                IP-Nets & 
                \makecell{learning rate: 0.0002 \\ batch size: 16 \\ hidden units: 128 \\ dropout: 0.1 \\ dropout recurrent: 0.0 \\ reconstruction fraction: 0.1} &
                \makecell{learning rate: 0.0005 \\ batch size: 32 \\ hidden units: 128 \\ dropout: 0.2 \\ dropout recurrent: 0.0 \\ reconstruction fraction: 0.5} & 
                \makecell{learning rate: 0.0002 \\ batch size: 64 \\ hidden units: 256 \\ dropout: 0.1 \\ dropout recurrent: 0.0 \\ reconstruction fraction: 0.2} & 
                \makecell{learning rate: 0.0005 \\ batch size: 32 \\ hidden units: 128 \\ dropout: 0.1 \\ dropout recurrent: 0.0 \\ reconstruction fraction: 0.5} & 
                \makecell{learning rate: 0.0005 \\ batch size: 64 \\ hidden units: 256 \\ dropout: 0.1 \\ dropout recurrent: 0.0 \\ reconstruction fraction: 0.2} 
                \\ \hline
                
                Transformer & 
                \makecell{learning rate: 0.0002 \\ batch size: 32 \\ num of layers: 3 \\ num of heads: 4 \\ hidden units: 512 \\ dropout: 0.2} & 
                \makecell{learning rate: 0.0002 \\ batch size: 64 \\ num of layers: 3 \\ num of heads: 8 \\ hidden units: 512 \\ dropout: 0.2} & 
                \makecell{learning rate: 0.0005 \\ batch size: 64 \\ num of layers: 3 \\ num of heads: 4 \\ hidden units: 512 \\ dropout: 0.2} & 
                \makecell{learning rate: 0.001 \\ batch size: 64 \\ num of layers: 3 \\ num of heads: 2 \\ hidden units: 256 \\ dropout: 0.1} & 
                \makecell{learning rate: 0.0005 \\ batch size: 32 \\ num of layers: 3 \\ num of heads: 8 \\ hidden units: 512 \\ dropout: 0.2} 
                \\ \hline
                
                SeFT & \makecell{
                learning rate: 0.001 \\ 
                batch size: 256 \\ 
                $\phi$ layers: 4 \\ 
                $\phi$ width: 128 \\ 
                $\phi$ dropout: 0.2 \\ 
                $\rho$ layers: 2 \\ 
                $\rho$ width: 512 \\ 
                $\rho$ dropout: 0.1   \\ 
                $\psi$ layer: 2 \\ 
                $\psi$ width: 64 \\ 
                dot prod dim: 128 \\ 
                attention dropout: 0.2 \\ 
                latent width: 128 \\
                max time scale: 1000 \\ 
                pos dims: 8} & 
                \makecell{
                learning rate: 0.0005 \\ 
                batch size: 128 \\ 
                $\phi$ layers: 2 \\ 
                $\phi$ width: 256 \\ 
                $\phi$ dropout: 0.2 \\ 
                $\rho$ layers: 3 \\ 
                $\rho$ width: 512 \\ 
                $\rho$ dropout: 0.5 \\ 
                $\psi$ layer: 2 \\ 
                $\psi$ width: 64 \\ 
                dot prod dim: 128 \\ 
                attention dropout: 0.1 \\ 
                latent width: 256 \\ 
                max time scale: 1000 \\ 
                pos dims: 8} & 
                \makecell{
                learning rate: 0.002 \\ 
                batch size: 256 \\
                $\phi$ layers: 4 \\ 
                $\phi$ width: 128 \\ 
                $\phi$ dropout: 0.1 \\ 
                $\rho$ layers: 2 \\ 
                $\rho$ width: 512 \\ 
                $\rho$ dropout: 0.3 \\ 
                $\psi$ layer: 2 \\ 
                $\psi$ width: 64 \\ 
                dot prod dim: 128 \\ 
                attention dropout: 0.1 \\ 
                latent width: 128 \\ 
                max time scale: 1000 \\ 
                pos dims: 8} & 
                \makecell{
                learning rate: 0.002 \\ 
                batch size: 256 \\ 
                $\phi$ layers: 4 \\ 
                $\phi$ width: 128 \\ 
                $\phi$ dropout: 0.1 \\ 
                $\rho$ layers: 2 \\ 
                $\rho$ width: 512 \\ 
                $\rho$ dropout: 0.3 \\ 
                $\psi$ layer: 2 \\ 
                $\psi$ width: 64 \\ 
                dot prod dim: 128 \\ 
                attention dropout: 0.1 \\ 
                latent width: 128 \\ 
                max time scale: 1000 \\ 
                pos dims: 8} & 
                \makecell{
                learning rate: 0.0005 \\ 
                batch size: 256 \\ 
                $\phi$ layers: 4 \\ 
                $\phi$ width: 128 \\ 
                $\phi$ dropout: 0.1 \\ 
                $\rho$ layers: 2 \\ 
                $\rho$ width: 512 \\ 
                $\rho$ dropout: 0.1 \\ 
                $\psi$ layer: 2 \\ 
                $\psi$ width: 64 \\ 
                dot prod dim: 128 \\ 
                attention dropout: 0.2 \\ 
                latent width: 128 \\ 
                max time scale: 1000 \\ 
                pos dims: 8}
                \\ \hline
                
                SAT-Transformer & \makecell{learning rate: 0.0002 \\ batch size: 32 \\ num of layers: 3 \\ num of heads: 8 \\ hidden units: 256 \\ dropout: 0.1 \\ learning rate2: x100} & \makecell{learning rate: 0.001 \\ batch size: 64 \\ num of layers: 2 \\ num of heads: 8 \\ hidden units: 256 \\ dropout: 0.1 \\ learning rate2: x20} &
                \makecell{learning rate: 0.0005 \\ batch size: 32 \\ num of layers: 3 \\ num of heads: 4 \\ hidden units: 512 \\ dropout: 0.1 \\ learning rate2: x20} &
                \makecell{learning rate: 0.0005 \\ batch size: 32 \\ num of layers: 3 \\ num of heads: 8 \\ hidden units: 512 \\ dropout: 0.2 \\ learning rate2: x50} &
                \makecell{learning rate: 0.0005 \\ batch size: 32 \\ num of layers: 3 \\ num of heads: 4 \\ hidden units: 512 \\ dropout: 0.2 \\ learning rate2: x20} 
                \\
                \toprule
            \end{tabular}
        }
        \caption{Selected hyperparameters for each model for each task}
    \label{table:a}
\end{table*}
    
\subsection{Baseline Comparison}
In this section, we compare the performances of the baseline models to other values in literature as shown in Table \ref{table:2}. The works done by \cite{horn2020set, che2018recurrent, shukla2019interpolation}, to our knowledge, show reliable performances of the same baseline networks. The evaluations done on PhysioNet or eICU data sets differ slightly from paper to paper. Due to this reason, the Table \ref{table:2} only shows performances evaluated on MIMIC-III data set. We have tuned the hyperparameters of the baseline models to achieve at least the performances shown in these works, if not better, for the MIMIC-III data set, putting the same effort into other data sets as well.

\renewcommand{\cellalign}{l}
\begin{table*}[h!]
    \centering
        \resizebox{17.2cm}{!}{
            \begin{tabular}{l l l l l}
                \toprule
                AUROC/AUPR &  Che et al. &  Shukla et al. & Horn et al. & Ours \\
                \hline
                GRU-Simple &
                \makecell{83.8 ± 0.8 / --} & 
                \makecell{-- / --} & 
                \makecell{82.8 ± 0.0 / 43.6 ± 0.4} & 
                \makecell{\textbf{85.4 ± 0.4 / 51.6 ± 0.6}} \\ 
                GRU-D &
                \makecell{85.3 ± 0.3 / --} & 
                \makecell{-- / --} & 
                \makecell{85.7 ± 0.2 / 52.0 ± 0.8} & 
                \makecell{\textbf{86.1 ± 0.7 / 52.8 ± 0.5}} \\ 
                IP-Nets &
                \makecell{-- / --} & 
                \makecell{\textbf{86.1 / 53.7}} & 
                \makecell{83.2 ± 0.5 / 48.3 ± 0.4} & 
                \makecell{85.4 ± 0.1 / 51.8 ± 0.9} \\ 
                Transformer &
                \makecell{-- / --} & 
                \makecell{-- / --} & 
                \makecell{82.1 ± 0.3 / 42.6 ± 1.0} & 
                \makecell{\textbf{84.8 ± 0.2 / 49.7 ± 0.3}} \\ 
                SeFT &
                \makecell{-- / --} & 
                \makecell{-- / --} & 
                \makecell{83.9 ± 0.4 / 46.3 ± 0.5} & 
                \makecell{\textbf{85.1 ± 0.3 / 46.2 ± 0.1}} \\ 
                \toprule
            \end{tabular}
        }
        \caption{Performances of the baseline models on MIMIC-III}
    \label{table:b}
\end{table*}

\end{document}